\documentclass[conference]{IEEEtran}

\IEEEoverridecommandlockouts                              

\usepackage{times}
\usepackage[numbers]{natbib}
\usepackage{graphics} 
\usepackage{mathtools}
\usepackage{graphicx}
\usepackage{tabularx}
\usepackage{caption}
\usepackage{subcaption}
\usepackage{wrapfig}
\usepackage{romannum}
\usepackage{amsmath,amssymb}
\usepackage{url}
\usepackage{bm}
\usepackage[table]{xcolor}
\usepackage[pagebackref=true,breaklinks=true,colorlinks=true,bookmarks=false,citecolor=blue]{hyperref} 

\usepackage{multirow}
\usepackage{booktabs}
\usepackage{pifont}

\usepackage[ruled,vlined]{algorithm2e}
\usepackage[noend]{algpseudocode}
\makeatletter
\let\OldStatex\Statex
\renewcommand{\Statex}[1][3]{%
  \setlength\@tempdima{\algorithmicindent}%
  \OldStatex\hskip\dimexpr#1\@tempdima\relax}
\renewcommand{\ALG@beginalgorithmic}{\normalsize}

\newcommand{\vx}{{\bf x}}

\newcommand{\vu}{{\bf u}}

\newcommand{\vR}{{\bf R}}

\newcommand{\vF}{{\bf F}}

\newcommand{\argmin}{\operatornamewithlimits{argmin}}

\DeclareCaptionFont{mysize}{\fontsize{8}{9.6}\selectfont}
\begin{document}

\title{
Safe Navigation in Unstructured Environments by Minimizing Uncertainty in Control and Perception
}

\author{Junwon Seo, Jungwi Mun, and Taekyung Kim \\
AI Autonomy Technology Center, Agency for Defense Development
\thanks{This work was supported by the Agency For Defense Development Grant funded by the Korean Government~(2023).}
}

\makeatletter
\makeatother

\maketitle


\begin{abstract}
Uncertainty in control and perception poses challenges for autonomous vehicle navigation in unstructured environments, leading to navigation failures and potential vehicle damage. This paper introduces a framework that minimizes control and perception uncertainty to ensure safe and reliable navigation. The framework consists of two uncertainty-aware models: a learning-based vehicle dynamics model and a self-supervised traversability estimation model. We train a vehicle dynamics model that can quantify the epistemic uncertainty of the model to perform active exploration, resulting in the efficient collection of training data and effective avoidance of uncertain state-action spaces. In addition, we employ meta-learning to train a traversability cost prediction network. The model can be trained with driving data from a variety of types of terrain, and it can online-adapt based on interaction experiences to reduce the aleatoric uncertainty. Integrating the dynamics model and traversability cost prediction model with a sampling-based model predictive controller allows for optimizing trajectories that avoid uncertain terrains and state-action spaces. Experimental results demonstrate that the proposed method reduces uncertainty in prediction and improves stability in autonomous vehicle navigation in unstructured environments.
\end{abstract}
\IEEEpeerreviewmaketitle


\section{Introduction}
Due to the complex and unpredictable nature of vehicle control and environmental perception, autonomous vehicle navigation in unstructured environments is challenging~\cite{borges2022survey}. Learning-based navigation methods for autonomous, high-speed vehicles in unstructured environments have demonstrated promising results in recent years. The success of fast-moving autonomous vehicles in off-road conditions can be attributed to two crucial factors. The first is learning-based vehicle control~\cite{williams_information_2017,lutter_deep_2019,  nagabandi_dexterous_2019,  yang_data_2019,shah2022gnm,kim_physics_2022}, and the second is the estimation of off-road terrain traversability~\cite{, kwak2017incremental_TE,ahtiainen2017normal,sevastopoulos2022survey,guan2022ga,contrastive_offroad,frey_2022locomotion,sathyamoorthy2022terrapn,TERP}.

Learning-based vehicle control has become crucial for autonomous vehicles since it facilitates optimal navigation performance while ensuring safety~\cite{kabzan_learning-based_2019, wu_daydreamer_2022, wang2021rough, kim_toast_2022}. Recent studies have demonstrated the effectiveness of combining learning-based vehicle dynamics models with model-based controllers~\cite{kahn2021badgr}. Also, estimating terrain traversability in off-road environments is essential for outdoor navigation. As off-road environments are rife with terrains that may generate uncertain vehicle reactions, failure to identify such uncertainty, which we refer to as \textit{perception failure}, could lead to navigation failure and even severe damage to the vehicle~\cite{Wellhausen_2020,pmlr-v155-ji21a,kerner2019novelty,ji2022proactive,seo2023learning}. Recent works present self-supervised methods for estimating traversability that take into vehicle-terrain interactions pertinent to navigation based on actual driving experiences~\cite{wellhausen2019should,Z_acoustic,Bekhti_verticala,Brooks_vibration,cai2022risk,gasparino2022wayfast,castro2022does}. They predict traversal costs derived from proprioceptive terrain interaction feedback using exteroceptive sensor measurements.

\begin{figure*}[t]
\centering
\includegraphics[width=0.9\textwidth]{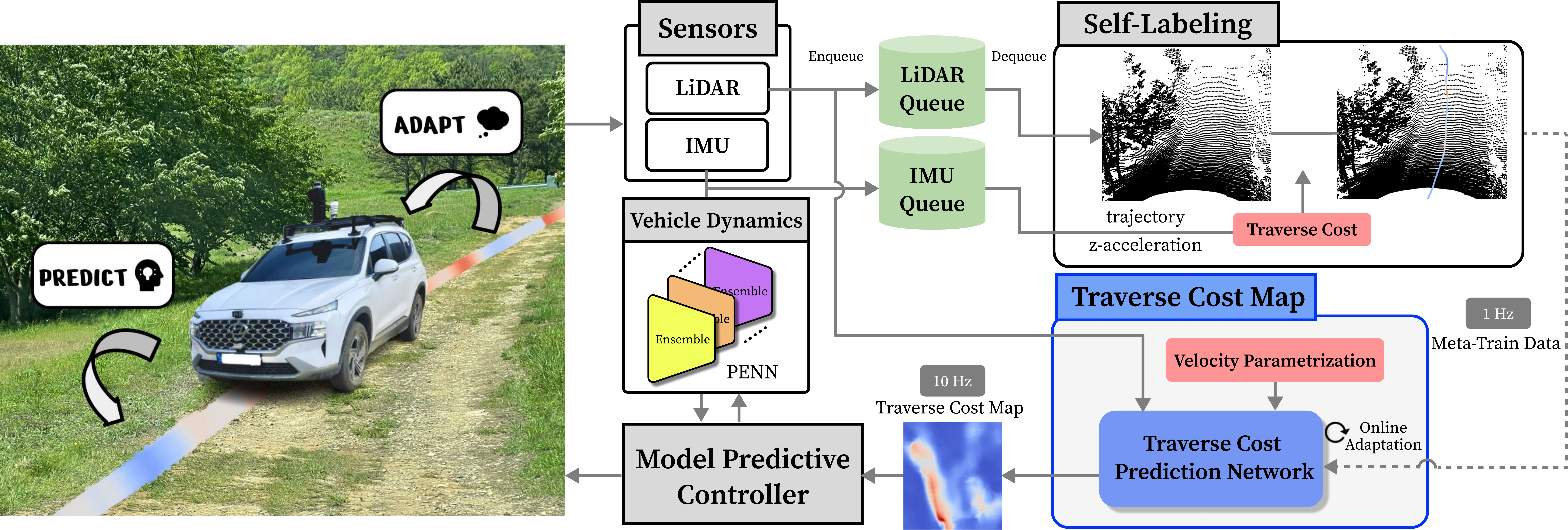}
\caption{Diagram of our navigation method that minimizes control and perception uncertainty. While training the dynamics model, active exploration is conducted, which minimizes epistemic uncertainty using a Probabilistic Ensemble Neural Network~(PENN). Our traversability cost prediction network reduces aleatoric uncertainty through online adaptation using self-labeled data. Our method enables safe and efficient navigation in unstructured environments by combining uncertainty-minimizing control and perception methods.}
\label{fig:concept}
\vspace*{-0.2in}
\end{figure*}

However, the primary limitation of these learning-based methods of control and perception is that the uncertainty of the model significantly affects the dependability of predictions~\cite{nguyen-tuong_model_2011,palazzo2020domain,fan2021learning,cai2022risk,seo_scate_2023}. Insufficient data to learn vehicle dynamics may result in uncertainty and suboptimal control performance~\cite{kahn_uncertainty-aware_2017,parsi_active_2020}. Self-supervised traversability estimation methods have inherent uncertainty problems since the interaction data only provides supervision for actually traversed regions~\cite{Wellhausen_2020,seo_scate_2023,frey2023fast}. Moreover, since the terrain traversability is predicted from a limited sensor configuration (e.g., a sparse LiDAR point cloud)~\cite{fei2021pillarsegnet,cheng2021s3cnet,han2021planning,peng2022mass,shaban2022semantic}, the estimation entails substantial aleatoric uncertainty.

For the safe and reliable navigation of autonomous vehicles in unstructured environments, it is necessary to minimize the uncertainty of predictions regarding both control and perception. The uncertainty-aware navigation can be performed through data exploration during training or online adaptation during deployment~\cite{nagabandilearning, BanerjeeHarrisonEtAl2020,RichardsAzizanEtAl2021, visca2022Meta}. If uncertainty cannot be resolved, reliable navigation can be enforced by preventing vehicles from accessing uncertain state-action spaces or terrain with high levels of uncertainty.

This paper proposes a framework for autonomous vehicle navigation in unstructured environments. To this end, we employ an uncertainty-aware learning-based dynamics model and a self-supervised traversability estimation model. The vehicle dynamics model is trained through a model-based reinforcement learning framework that quantifies epistemic uncertainty. After learning the dynamics model, we generate an uncertainty-aware traversability cost map. Using the self-supervised traversability cost of terrains derived from interaction experiences, the traversability model is capable of online adaptation to minimize uncertainty. We integrate the dynamics model and traversability cost map with a sampling-based model predictive controller. The controller allows for safe navigation by optimizing the trajectory to avoid regions and actions with high uncertainty.

\section{Methods}
\subsection{Uncertainty Minimization in Control}
We aim to learn a vehicle dynamics model, $\vF$, which can minimize uncertainty using active exploration. The system dynamics can be defined as ${\vx}_{t+1} = \vF(\vx_t, \vu_t)$ where $\vF$ is a nonlinear function, $\vx_t \in \mathbb{R}^{n}$ and $\vu_t \in \mathbb{R}^{m}$ are the observed state vector and applied action input at time~$t$, respectively. Following our previous work~\cite{kim2023bridging}, we employ Probabilistic Ensemble Neural Network~(PENN) to approximate the vehicle dynamics ~$\vF$~\cite{chua_deep_2018, buckman_sample-efficient_2018}.

The disagreement between the ensemble model of a state-action pair~$D(\vx_t, \vu_t)$ can be utilized to encourage active exploration in order to collect additional training data. Using a closed-form Jensen-Rényi Divergence~\cite{renyi_measures_1961, wang_closed-form_2009, shyam_model-based_2019} that can be calculated in real-time, our previous method~\cite{kim2023bridging} demonstrated that the epistemic uncertainty of a given state action pair can be effectively quantified. During the training phase, the vehicle is instructed to collect data with high disagreement so as to accumulate data with high epistemic uncertainty. As additional training data are collected, ensemble models will gradually converge toward increasingly similar predictions, resulting in an eventual reduction in model uncertainty.

For active exploration, Model Predictive Path Integral (MPPI)~\cite{williams_information_2017} control is utilized. MPPI is the state-of-the-art sampling-based MPC whose parallelizable structure enables real-time implementation using Graphic Processing Units (GPUs)~\cite{williams_information-theoretic_2018}. Denoting by $U = \{\vu_0, \dots, \vu_{T-1}\}$ with a fixed time horizon~$T$, the MPPI algorithm seeks an optimal control sequence~$U^{*}$ such that:
\begin{equation}
    U^{*} = \underset{U}{\operatorname{argmin}} \, \mathbb{E} \left[\phi(\vx_{T})+\sum_{t=0}^{T-1} \mathcal{L}\left(\vx_t, \vu_t\right) \right] ,
\end{equation}
where $\phi\left( \cdot \right)$ is a state-dependent terminal cost. 

For the active exploration, the model disagreement is added to the controller's objective in order to jointly achieve task-dependent cost optimization and active exploration:
\begin{equation}
    \mathcal{L}^{\text{active}}\left(\vx_t, \vu_t\right) = q \left( \vx_{t} \right) + \frac{1}{2} \vu_{t}^{\top} \vR \vu_{t} - w_{D} \, D(\vx_t, \vu_t) \,,
    \label{eq:jrd_cost}
\end{equation}
, where $w_{D} > 0$ is a weighting constant and $q \left( \vx_{t} \right)$ is an arbitrary state-dependent running cost function . The vehicle is encouraged to select behaviors that result in the exploration of ambiguous state-action spaces of the dynamic model.

\subsection{Uncertainty Minimization in Perception}
\subsubsection{Anomaly Detection for Minimizing Epistemic Uncertainty}\label{sec:uncertainty}

For effective navigation in off-road environments, the traversability of terrain can be determined by interactions between terrain and a vehicle~\cite{wellhausen2019should,Z_acoustic}. However, data on such interactions is restricted to traversed regions only, and data collection in non-traversable regions is not feasible. This nature produces overconfident predictions, resulting in navigational failure or even catastrophic outcomes~\cite{Wellhausen_2020}.

This issue can be mitigated by employing methods for detecting anomalies~\cite{kerner2019novelty,Wellhausen_2020,pmlr-v155-ji21a}. Regions that a vehicle interacted with are designated as positive, on the assumption that terrains with similar geometric properties to those regions are normal and thus less uncertain. Then, we train a binary classifier capable of identifying points that are dissimilar to positive points, indicating high epistemic uncertainty. These regions are designated for avoidance during navigation. The method is described in detail in our previous work~\cite{seo_scate_2023, seo2023learning}.

\subsubsection{Uncertainty-Aware Traversability Cost Map}\label{sec:costmap}

Off-road environments are fraught with bumps and obstacles of varying shapes, despite being in the same terrain class. For safe and effective navigation of a vehicle in off-road terrain, estimating the nuanced traverse cost of the terrain is necessary. A path planner can optimize a trajectory that minimizes disturbances during navigation with the predicted cost. Therefore, we generate dense and continuous traverse cost maps in bird's eye view~(BEV) from a single LiDAR point cloud.

The z-axis linear acceleration measured by an Inertial Measurement Unit~(IMU) is used to define the traversability that can stabilize navigation, as this component effectively communicates the terrain's bumpiness~\cite{Bekhti_verticala}. In addition, because our controller employs a vehicle dynamics model that is ignorant of vertical motions~\cite{kim_smooth_2022, kim_physics_2022}, the definition of traversability based on vertical acceleration can be helpful in preventing control performance losses. Motivated by recent work~\cite{castro2022does}, we define traverse cost using the spectral power of z-acceleration produced by a continuous wavelet transformation. 

\begin{figure}[t]
\centering
\includegraphics[width=1.0\linewidth]{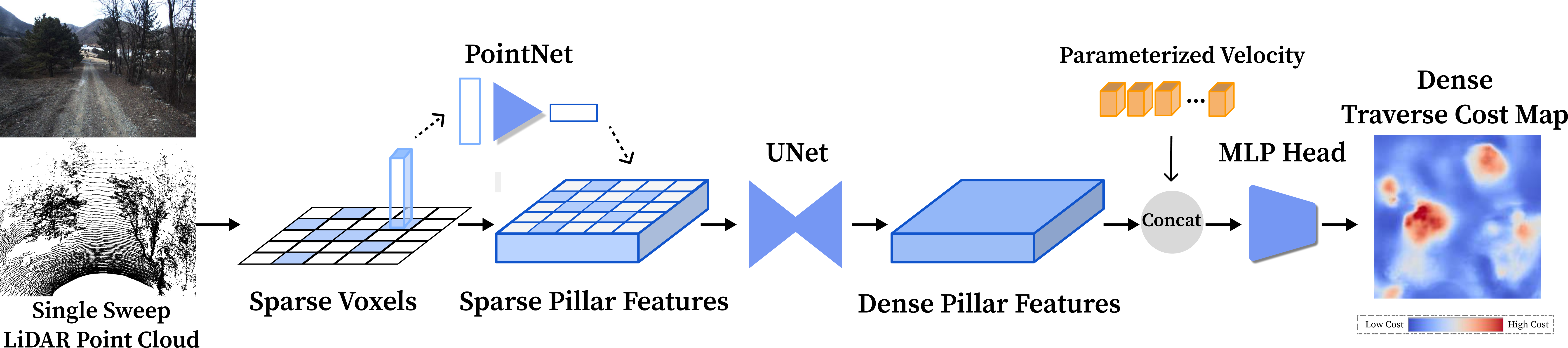}
\caption{Overview pipeline of the traversability cost prediction network.}
\label{fig:pipeline}
\vspace*{-0.25in}
\end{figure}

The network is trained to produce a dense traverse cost map using a sparse single-sweep LiDAR point cloud in BEV, as shown in Fig.~\ref{fig:pipeline}. Following PointPillars~\cite{lang2019pointpillars}, each point is discretized into sparse voxels with a grid size of 0.2m. Using PointNet~\cite{qi2017pointnet}, each voxel is converted into sparse pillar features. Then, a U-Net~\cite{unet} structured network, which has an encoder-decoder architecture with skip connections, is employed to generate a dense pillar feature map. It progressively reduces the spatial size of features and captures higher-level semantic information, while the decoder upsamples feature maps to recover spatial information. Then, the dense pillar features are then concatenated with parameterized velocity to produce velocity-conditioned cost maps. The velocity is parameterized with Fourier feature mapping~\cite{tancik2020fourier} to incorporate vehicle's velocity into the costmap prediction~\cite{castro2022does}.

Finally, the MLP head predicts the mean ${\mu}_i$ and standard deviation ${\sigma}_i$ of the traversability for each pillar $i$. The network is trained to minimize the Gaussian log-likelihood:
\begin{equation}\label{gaussiannll}
    \mathcal{L}^{\text{traverse}}\left(\tau, \bm{\theta}\right) = \frac{1}{2}\left(\log({\sigma}_i) +  \frac{({\mu}_i - c_i)^2}{{\sigma}_i} \right),
\end{equation}where $\tau$, $c_i$, and $\bm{\theta}$ represent a trajectory, the cost of the the trajectory that is associated with the pillar $i$, and model parameter, respectively.

\subsubsection{Online Adaptation for Minimizing Aleatoric Uncertainty}\label{sec:meta}

\begin{algorithm}[h]
\footnotesize
\SetKwInOut{Input}{Given}
\Input{$\bm{\theta}$:
$\mathcal{D}$: Interaction data from various environments\;
$M, K$: Number of past and future timesteps\;
$N$: Number of sampled trajectories\;
$N_A$: Number of the inner gradient steps\;
$\alpha, \beta$: Learning rate for inner and outer gradient steps\;
Randomly initialize $\bm{\theta}$\;
\For{$i \leftarrow 0$ \KwTo \text{maximum iterations}}{
    \For{$j \leftarrow 0$ \KwTo $N-1$}{
        Sample $\tau(t-M, t)$, $\tau(t, t+K)$ $\sim \mathcal{D}$\;
        Self-Label $\tau(t-M, t)$ and $\tau(t, t+K)$\;
        $\bm{\theta}' \leftarrow \bm{\theta}$\;
        \For{$k \leftarrow 0$ \KwTo $N_A-1$}{
            $\bm{\theta}' \leftarrow \bm{\theta}' - \alpha \nabla_{\bm{\theta'}} \mathcal{L}^{\text{traverse}}\left(\tau(t-M, t), \bm{\theta'}\right)$\;
        }
        $\mathcal{L}_j \leftarrow \mathcal{L}^{\text{traverse}}\left(\tau(t, t+K), \bm{\theta'}\right)$
    }
    $\bm{\theta} \leftarrow  \bm{\theta} - \beta \nabla_{\bm{\theta}} \frac{1}{N} \sum\limits_{j=1}^N \mathcal{L}_j$
}
}
\caption{Meta Learning of Traverse Cost \label{Algorithm:meta}} 
\end{algorithm}

A significant degree of aleatoric uncertainty occurs when learning traverse cost from a single-sweep LiDAR point cloud. In real-world off-road environments, there are a variety of terrain types, and the terrain and vehicle properties that influence traversability (e.g., deformability, friction, and roughness) differ according to a number of factors. However, such subtle differences cannot be precisely captured by a LiDAR point cloud. The ground-truth traverse cost of terrain captured in different environments would vary considerably, while the geometric characteristics of terrain would be comparable. Learning a global traversability model using a large dataset~$\mathcal{D}$ of multiple environments leads to high aleatoric uncertainty in estimation.

To resolve this issue, we propose a model that can effectively learn a global traversability model that is capable of quickly adapting to a new environment based on its recent experiences. We use Model-Agnostic Meta-Learning~(MAML)~\cite{finn2017model} for online adaptation of the traversability model. It aims to find the initial parameters of the network so that adaptation with a few gradient descent steps from this initialization leads to effective generalization to current circumstances. By doing so, the model can incorporate driving data collected in various environments for learning traverse cost from geometric properties without confusion.

While terrain properties vary significantly in different environments, we assume that the environment is locally consistent. Consequently, each segment of a trajectory is regarded as a separate \textit{task}, denoted as $\tau$. The traversability model is trained to adapt using the meta-train data of $M$ past timesteps, $\tau\left(t-M,t\right)$, to predict the traverse cost of the next $K$ timesteps, $\tau\left(t,t+K\right)$, as follows: 
\begin{equation}\label{eq:meta}
    \begin{aligned}
    \argmin_{\bm{\theta}} \hspace{10pt} & \mathbb{E}_{\tau\left(t-M, t+K\right) \sim \mathcal{D}}  \big[ \mathcal{L}^{\text{traverse}}\left(\tau(t, t+K), \bm{\theta}'\right)\big] \\   
    & \text{s.t.:} \hspace{10pt} \bm{\theta}' = \bm{\theta} - \alpha \nabla_{\bm{\theta}} \mathcal{L}^{\text{traverse}}\left(\tau(t-M, t), \bm{\theta}\right).
    \end{aligned}
\end{equation} The past $M$ timesteps provide insight on how to adapt the model to precisely predict future trajectories' traverse costs. As illustrated in Fig.\ref{fig:concept}, the model can online-adapt during the deployment phase using automatically generated meta-train data in a self-supervised manner. Algorithm~\ref{Algorithm:meta} describes the meta-learning-based training procedure for the global traversability model. The trained global model is applicable in a variety of environments and is even able to adjust to unseen terrain. 

\begin{table}[b]
\vspace*{-0.1in}
\centering
\renewcommand {\arraystretch}{1.1}
\caption{Validation error of traversability estimation models for real-world driving data. Our method shows a significant margin compared to the baseline.}
\footnotesize
\label{tab:quantitative}
\resizebox{1.0\linewidth}{!}{%
    \begin{tabular}{cccccc}
        \toprule
        & \bf{Unpaved} & \bf{Rough} & \bf{Grassland} & \bf{Profiled Road} & \bf{Simulation} \\
        \midrule \midrule
        
        \multicolumn{1}{c|}{\textit{Baseline}}      & 0.1222 & 0.0196 & 0.2460 & 0.2039 & 0.7263 \\
        \multicolumn{1}{c|}{\textit{w.o. Adaptation}} & 0.0713 & 0.0171 & 0.1961 & 0.1907  & 0.6668 \\
        \multicolumn{1}{c|}{\textit{w. Adaptation}} & \textbf{0.0114} & \textbf{0.0015} & \textbf{0.1767} & \textbf{0.1523} & \textbf{0.5725} \\
\end{tabular}%
}
\end{table}

\section{Experiments \label{sec:exp}}
\begin{figure*}[t]
\centering
\includegraphics[width=0.95\linewidth]{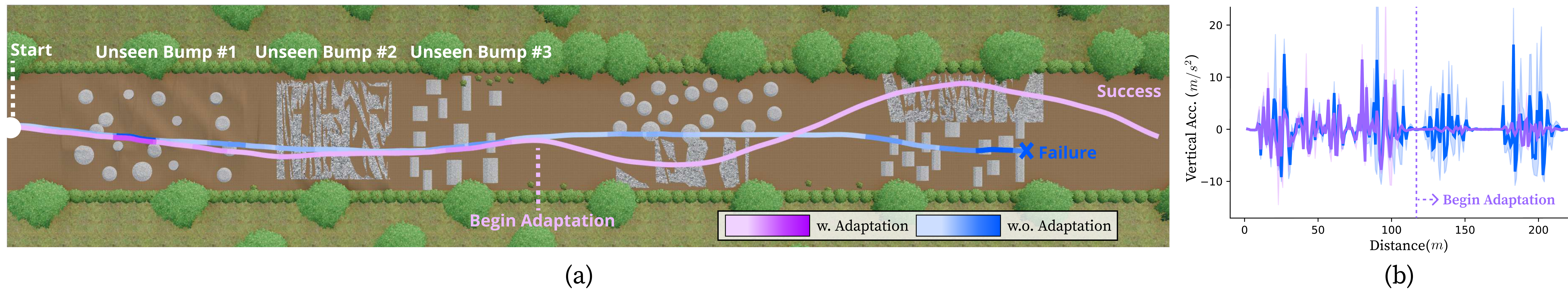}
\caption{(a) The off-road race track designed for online adaptation~(\textbf{Q3}). The vehicle trajectories are visualized, and the colors of the lines illustrate the rotational impacts exerted on the vehicle. (b) Vertical acceleration of the vehicle during navigation. By conducting online adaptation, the vehicle can plan paths that can minimize impacts exerted on the vehicle, leading to stable navigation in off-road.
}
\label{fig:adaptation}
\vspace*{-0.15in}
\end{figure*}

Our previous works have demonstrated our methods for minimizing epistemic uncertainty in vehicle dynamics and terrain traversability~\cite{kim2023bridging,seo_scate_2023}. Therefore, we concentrate experiments on the integration of methods for autonomous navigation, particularly in terms of online adaptation for traversability~(see Section~\ref{sec:meta}). Our experiments address the following key questions: \textbf{(Q1)} Can our method reduce aleatoric uncertainty for learning traversability cost? \textbf{(Q2)} Is our method helpful for safe and stable navigation in unstructured environments? \textbf{(Q3)} Can our traversability estimation model online adapt effectively to unknown terrains for navigation?

\subsection{Minimizing Aleatoric Uncertainty of Traversability Cost}

\textbf{Experimental Setup}
We quantitatively evaluate our meta-learning method for traversability cost prediction~(\textbf{Q1}). We collect driving data in various types of terrain with our vehicle~(See Fig.~\ref{fig:concept}) equipped with OS1-128 LiDAR and IMU. Also, the driving data is obtained in a simulation environment consisting of randomly patterned rough terrain and bumps. Approximately five hours of driving data are utilized to train the network. For comparison, a global model~(\textit{Baseline}) is simply trained without adaptation.

\textbf{Experimental Result}
The Table~\ref{tab:quantitative} provides the Mean Square Error~(MSE) between the ground-truth traversability and the predicted traversability cost. In every category, our method produces superior results compared to the baseline. It means the non-meta-learned baseline failed to converge due to high aleatoric uncertainty in ground-truth traversability gathered in various terrains, whereas our meta-learned model can converge well by incorporating such uncertainty during training. Our method even shows better performance without adaptation during the inference~(\textit{w.o. Adaptation}), meaning that it finds a better initial parameter. Moreover, the performance improves as the model adapts using the meta-train data derived from past interaction experiences~(\textit{w. Adaptation}). 
\subsection{Safe Navigation in Unstructured Environment}

\textbf{Experimental Setup}
We evaluate \textbf{Q2}-\textbf{Q3} using a high-fidelity vehicle dynamics simulator~-~IPG CarMaker. For all experiments, navigation is performed using only local traversability maps generated from LiDAR point clouds, with no prior knowledge or global map of the environments. For each method, navigation is performed $15$ times.

\begin{figure}[ht]
\centering
\includegraphics[width=0.9\linewidth]{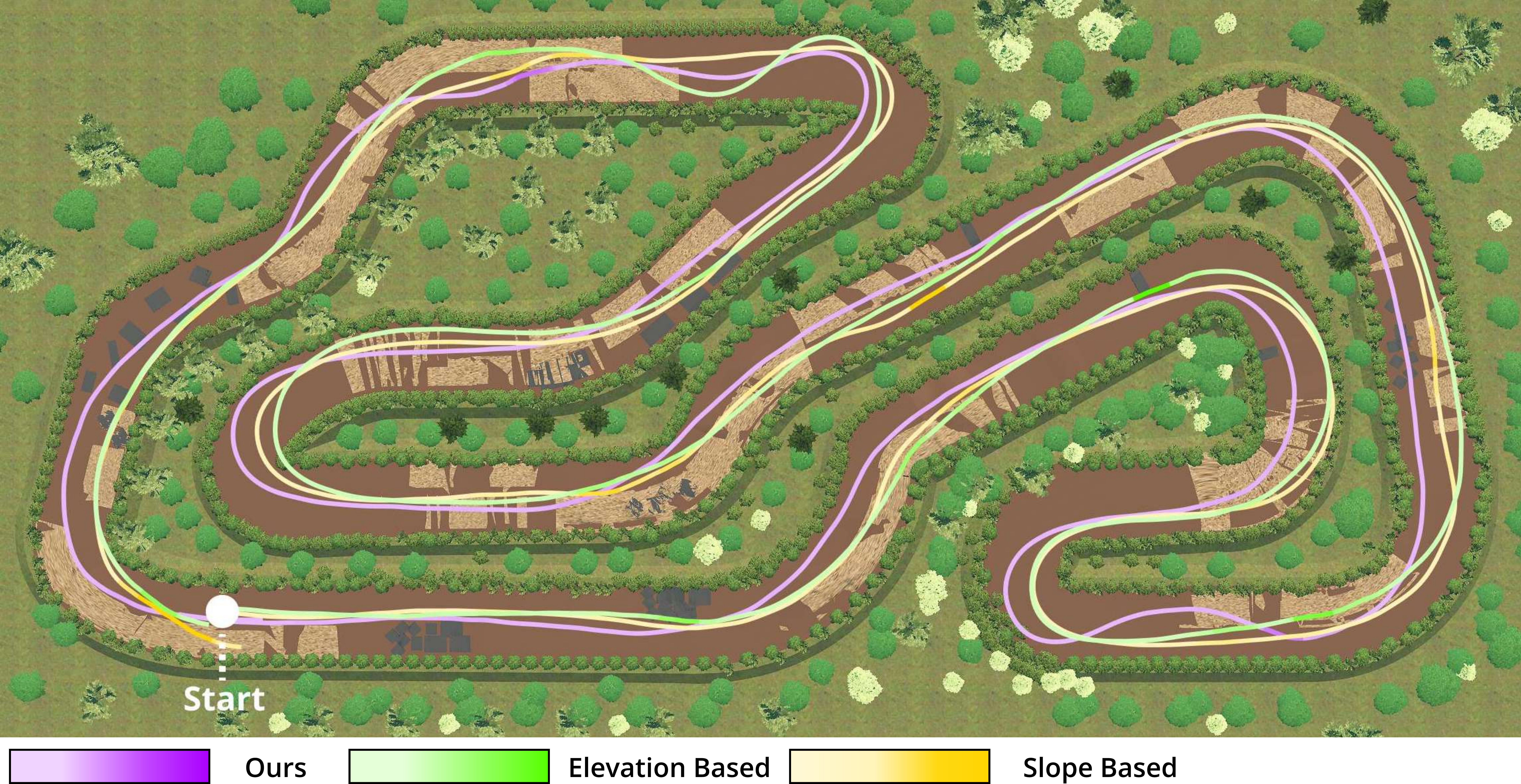}
\caption{The off-road race track designed in the IPG CarMaker simulator for navigation experiments. The vehicle trajectory is displayed.}
\label{fig:driving}
\vspace*{-0.25in}
\end{figure}

A realistic off-road environment is designed with large bumps and randomly patterned rough terrain (see Fig.~\ref{fig:driving})to conduct navigation with traversability maps produced by various methods~(\textbf{Q2}). Traversability maps are produced from our method with and without online adaptation, an elevation-map based method~(\textit{Elevation Based})~\cite{fankhauser2014robot,miki2022elevation}, and a slope-based method~(\textit{Slope Based})~\cite{sock2016probabilistic,kim2017traversable}.

\begin{table}[b]
\vspace*{-0.1in}
\centering
\small
\renewcommand {\arraystretch}{1.15}
\caption{The average and maximum motions of the vehicle across all trials.}
\resizebox{1.0\linewidth}{!}{
\label{tab:results}
    \begin{tabular}{cccccccccc}
        \toprule
        \multirow{2}{*}{\textbf{Method}} & \multirow{2}{*}{\textbf{Success Rate}} & \multicolumn{2}{c}{\bf{Vertical Vel. [m/s]}} & \multicolumn{2}{c}{\bf{Vertical Acc. [m/s$^\text{2}$]}} & 
        \multicolumn{2}{c}{\bf{Roll Acc. [rad/s$^\text{2}$]}} &
        \multicolumn{2}{c}{\bf{Pitch Acc. [rad/s$^\text{2}$]}} \\
        \cmidrule(l{4pt}r{4pt}){3-4} \cmidrule(l{4pt}r{4pt}){5-6} \cmidrule(l{4pt}r{4pt}){7-8} \cmidrule(l{4pt}r{4pt}){9-10} & & \textbf{Mean} & \textbf{Max} & \textbf{Mean} & \textbf{Max} & \textbf{Mean} & \textbf{Max} & \textbf{Mean} & \textbf{Max}\\
        \midrule
        Elevation Based & 7 / 15 & 0.152 & 2.766 & 1.254 & 40.108 & 1.287 & 54.783 & 1.177 & 34.029\\
        Slope Based & 6 / 15 & 0.128 & 1.846 & 0.999 & 30.481 & 1.018 & 38.910 & 0.953 & 26.822\\
        Ours(w.o. Adaptation) & 14 / 15 & 0.109 & 1.690 &  0.868 & 27.948 & 0.899 & 36.022 & 0.878 & 27.046 \\
        \rowcolor[HTML]{C0C0C0}Ours(w. Adaptation) & 15 / 15 & 0.106 & 1.591 & 0.854 & 25.686 & 0.892 & 32.217 & 0.849 & 22.079\\
        \bottomrule
    \end{tabular}
    }
\label{table:safe}
\end{table}

An additional off-road environment is designed with several unknown types of bumps (see Fig.~\ref{fig:adaptation}), to validate that our model is capable of online-adapting to unknown terrains~(\textbf{Q3}), which eventually results in safe navigation. The vehicle begins to adapt after experiencing three unseen types of bumps. This allows the vehicle to effectively optimize trajectories that minimize perturbation when re-encountering the bumps.

Based on our prior work \cite{kim_smooth_2022}, we formulate a simple running cost function $q(\vx_t)$ of the controller, which is designed to minimize the traversability cost of a trajectory while maintaining the $30$ km/h target speed:
\begin{equation}
    q(\vx_t) = \alpha_1{\text{Track}(\vx_t)} + \alpha_2{\text{Stable}(\vx_t)} + \alpha_3{\text{Speed}(\vx_t)},
\end{equation} where $\text{Track}(\vx_t)$ and $\text{Stable}(\vx_t)$ are assigned based on the local traversability map. $\text{Track}(\vx_t)$ imposes a significant penalty based on (Sec.~\ref{sec:uncertainty}) and $\text{Stable}(\vx_t)$ is based on the traversability cost map generated by our method (Sec.~\ref{sec:costmap}).

\textbf{Experimental Results} The trajectories taken during navigation are shown in Fig.~\ref{fig:driving}. Compared to other rule-based methods, the vehicle using our traversability map navigates along trajectories that minimize disturbances. Table.~\ref{table:safe} shows the average values of the vehicle's vertical and angular motions. It shows that our map of traversability can assist in stabilizing the vehicle during off-road navigation.

Fig.~\ref{fig:adaptation} shows the navigation results in the scene designed for evaluating online-adaptation. The vehicle experiences unseen bumps and effectively adapts the model to incorporate the experience. After adaptation, the controller determines to avoid difficult bumps, whereas the vehicle without adaptation fails to avoid them, resulting in a rollover. By beginning to online-adapt the network, the vehicle is capable of reducing the impact exerted on it, while the vehicle not conducting adaptation continues to experience enormous impacts.

\addtolength{\textheight}{0 cm}   



\hypersetup{
    urlcolor = .,
}

\bibliographystyle{IEEEtran}
\bibliography{mybib.bib}

\end{document}